\documentclass{article}

\usepackage{arxiv}

\usepackage[utf8]{inputenc} 
\usepackage[T1]{fontenc}    
\usepackage{hyperref}       
\usepackage{url}            
\usepackage{booktabs}       
\usepackage{amsfonts}       
\usepackage{nicefrac}       
\usepackage{microtype}      
\usepackage{lipsum}		
\usepackage{graphicx}
\usepackage[numbers]{natbib}
\usepackage{doi}
\usepackage{color}
\usepackage{subcaption,caption,algorithm,algorithmicx,amsmath,enumitem,cleveref}

\title{PCIM: Learning Pixel Attributions via Pixel-wise Channel Isolation Mixing in High Content Imaging}

\author{\hspace{1mm}Daniel Siegismund\\
Genedata AG\\
Margarethenstrasse 38\\
4053 Basel \\
Switzerland\\
\texttt{daniel.siegismund@genedata.com} \\
\And
\hspace{1mm}Mario Wieser\\
Genedata AG\\
Margarethenstrasse 38\\
4053 Basel \\
Switzerland\\
\texttt{mario.wieser@genedata.com} \\
\And
\hspace{1mm}Stephan Heyse\\
Genedata AG\\
Margarethenstrasse 38\\
4053 Basel \\
Switzerland\\
\texttt{stephan.heyse@genedata.com} \\
\And
\hspace{1mm}Stephan Steigele\\
Genedata AG\\
Margarethenstrasse 38\\
4053 Basel \\
Switzerland\\
\texttt{stephan.steigele@genedata.com}\\
}

\renewcommand{\shorttitle}{PCIM}

\hypersetup{
    pdftitle={PCIM: Learning Pixel Attributions via Pixel-wise Channel Isolation Mixing in High Content Imaging},
    pdfsubject={q-bio.NC, q-bio.QM},
    pdfauthor={D. Siegismund},
    pdfkeywords={Biomedical Imaging, Interpretable Machine Learning,Explainable AI, Image Pixel Importance, Feature Importance
    Attribution},
}

\begin{document}
    \maketitle

    \begin{abstract}
        Deep Neural Networks (DNNs) have shown remarkable success in various computer vision tasks. However, their black-box nature often leads to difficulty in interpreting their decisions,
        creating an unfilled need for methods to explain the decisions, and ultimately forming a barrier to their wide acceptance especially in biomedical applications. This work introduces a
        novel method, Pixel-wise Channel Isolation Mixing (PCIM), to calculate pixel attribution maps, highlighting the image parts most crucial for a classification decision but without the
        need to extract internal network states or gradients.
        Unlike existing methods, PCIM treats each pixel as a distinct input channel and trains a blending layer to mix these pixels, reflecting specific classifications.
        This unique approach allows the generation of pixel attribution maps for each image, but agnostic to the choice of the underlying classification network.
        Benchmark testing on three application relevant, diverse high content Imaging datasets show state-of-the-art performance,
        particularly for model fidelity and localization ability in both, fluorescence and bright field High Content Imaging. PCIM contributes as a unique and effective method for
        creating pixel-level attribution maps from arbitrary DNNs, enabling interpretability and trust.
    \end{abstract}

    \keywords{Biomedical Imaging, Interpretable Machine Learning,Explainable AI, Image Pixel Importance, Feature Importance
    Attribution}

    \section{Introduction}
    In recent years, deep neural network architectures (DNNs) demonstrated superior performance in a large variety of computer vision tasks such as image classification or segmentation (\cite{Kirillov_2023_ICCV,NIPS2012_c399862d}).
    Despite their success compared to hand-designed features such as Sift (\cite{Lowe:2004:DIF:993451.996342}), DNN come with the drawback of a black-box function which is challenging to interpret the prediction outcome.
    Especially in settings that demand high trust in the decisions such as medical and pharmacological application areas DNNs suffer from a lack of trust towards their decisions (
    \cite{ribeiro2016should,lipton2018mythos,rao2024better}).

    Recent research on the interpretability of DNNs primarily focuses on the significance of image pixels in natural images. Therefore various methods ranging from gradient-based approaches,
    perturbation techniques, to exemplar-based methods have been developed to understand importance of an image pixel with respect to the prediction result (\cite{nazir2023survey}). In addition, there are three further lines of research focusing on interpretability of DNNs: Activation-based, backpropagation-based and perturbation-based approaches. Activation-based methods(\cite{zhou2016learning}) are using weight activation maps that are weighted by their gradients (Grad-CAM \cite{selvaraju2016grad},
    Grad-CAM++ \cite{chattopadhay2018grad}) or importance scores which are based on classification performance (\cite{ramaswamy2020ablation}). In contrast, backpropagation-based methods calculate gradients of the class score against each pixel to determine the importance of individual pixels (\cite{simonyan2013deep,shrikumar2017learning,sundararajan2017axiomatic}).
    Perturbation-based methods alter inputs to assess pixel importance, using occlusion (\cite{petsiuk2018rise}) and mask adjustment (\cite{fong2017interpretable}).
    
    The pixel importance determination is useful and valuable not only for natural images but especially for biomedical imaging where Grad-CAM(++) is most frequently applied
    (\cite{nazir2023survey,van2022explainable}). Here, pixel attribution maps assessing the pixel importance play a key role in interpretation,
    assisting to understand whether the biologically relevant image parts are reflected in the decision-making process \cite{nazir2023survey,rohrl2023explainable}, e.g. when deciding to take a certain drug candidate forward in pharma research.

    In particular, High content imaging (HCI) performed through automated microscopy in large volumes, heavily benefit from highlighting predictive image areas
    via attribution maps. The reason is that HCI captures biological
    activity within cells or tissues following the interference with specific agents
    (such as compounds, antibodies, or siRNAs) using fluorescence or bright-field microscopy (\cite{buchser2014assay,usaj2016high}) where such attribution maps give the scientist insights about the experimental outcome.

    In this work, we propose a novel approach called PCIM (pixel-wise channel isolation mixing) for calculating pixel attribution maps based on deep image channel mixing, as introduced in (\cite{siegismund2023learning}).
    Here, we treat each pixel as a distinct input channel and train a corresponding blending layer which mixes the pixels to reflect specific classification outcomes. More specifically, the weights of the underlying classification network remain fixed, while the trained mixing weights of the blending layer
    dynamically capture the pixel importance. The main advantage of this approach is that the process is model-agnostic of the underlying DNN architecture for generating attribution maps.

    In order to evaluate PCIM, the method is compared to multiple baseline approaches representing the variety of available methods to achieve attribution
    maps, relying on VGG16 (\cite{simonyan2014very}) and EfficientNetV2 (\cite{tan2021efficientnetv2}) classification networks. As evaluation datasets, two pharma-industry relevant High Content image datasets are used \cite{peddibhotla2013discovery,ljosa2012annotated}.
    The main achievements of the presented study are following: (1) We introduce a novel method (PCIM) to create pixel-level attribution maps via channel isolation mixing from arbitrary Deep Neural Networks (DNNs),
    avoiding the need to use internal network states or gradients. PCIM is unique in its approach to creating pixel attribution maps, setting it apart from other methods that have been previously published (2) We benchmark PCIM on three different
    real-world high content imaging datasets: NTR1 (fluorescence microscopy) (\cite{peddibhotla2013discovery}), BBBC054 (bright-field microscopy) (\cite{ljosa2012annotated}) and BBBC010 (fluorescence microscopy) (\cite{ljosa2012annotated}) which achieves
    state-of-the-art performance against the baseline method. (3) We qualitatively explore how the interplay between pixel attributions from PCIM and the underlying biology of the different assays significantly enhances our understanding of DNN classifications.

    \section{Materials and Methods}
    \label{sec:methods}
    \subsection{PCIM Idea}
\begin{figure}[h]
    	\centering
        \includegraphics[width=0.85\linewidth]{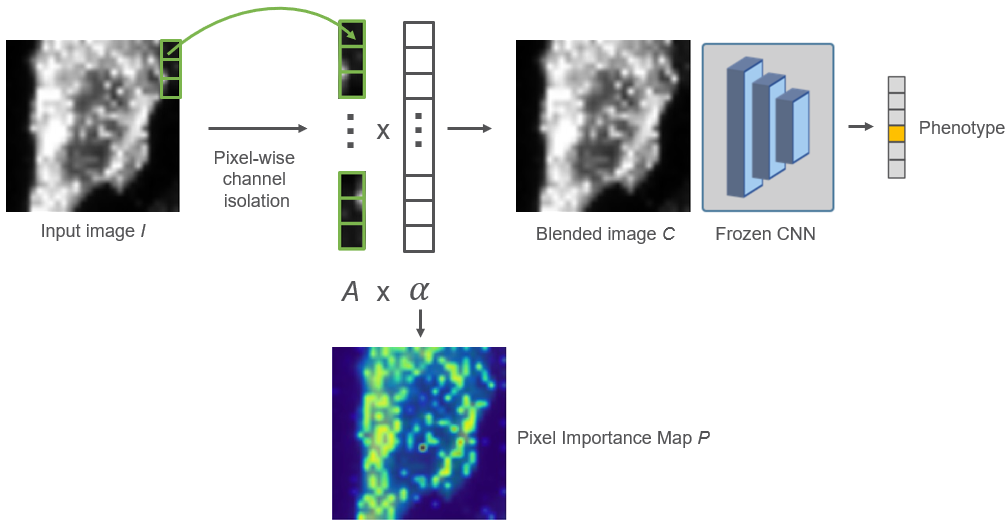}
        \caption{Pixel-Wise Channel Isolation Mixing (PCIM). 1. Transformation of the image pixels (green boxes) into an isolated channel vector $A$. 2. The vector is mixed with an importance vector $\alpha$ into the resulting blended image $C$, fed into the frozen CNN. 3. Following the training, extract the pixel importance map $P$ from the importance vector $\alpha$.}
        \label{fig:method_overview}
    \end{figure}

    We assume as a pre-requisite a pre-trained image classification network with frozen weights in the subsequent step. 
    PCIM performs three steps to obtain the pixel attribution maps. 
    \begin{enumerate}[label=\textbf{\arabic*})]
        \item \textbf{Pixel Isolation:} Each pixel in an input image is isolated into one independent channel. Then we consider just the information related to the specific pixel.
        \item \textbf{Pixel Channel Mixing:} Train a mixing layer that combines the isolated pixels from the different channels, while setting high importance on pixels which show a high contribution to the classification result.
        \item \textbf{Pixel Importance Map:} After training, we extract the pixel importance map for a certain image from weights of the mixing layer.
    \end{enumerate}
    For a graphical illustration of PCIM please refer to Figure \ref{fig:method_overview}.

    \subsubsection{PCIM Algorithm}
    We use for PCIM the channel importance mixing approach as introduced in \cite{siegismund2023learning}. Contrasting the original approach, each image pixel is taken as a separate channel while \cite{siegismund2023learning} is using the image channels itself. The algorithm starts by processing a single-channel input image $I \in \mathbb{R}^{h \times w}$ where $h$ denotes the height, $w$ the width and $p=h \times w$ the number of pixels in the image. Subsequently, the image $I$ is split into a vector $A^{p \times 1}$ with $p$ distinct channels, each holding only one pixel value on the spatial position $i,j$ which is processed in the mixing layer.
    Following \cite{zhang_deep_2020,kokalj_why_2019,siegismund2023learning}, we make use of trainable $\alpha \in  \mathbb{R}^{p \times 1}$ values as the weights for each channel to obtain the blended image $C$:

    \begin{equation}
        C = \sum^p_i \alpha_i \cdot A_i
        \text{ where:   }
        \alpha_i \geq 0,
        \label{eq:P}
    \end{equation}
    
    where $\alpha_i$ is multiplied with each pixel channel $A_i$ and the parameter $p$ defines the number of pixels. Following, the resulting vector $C \in  \mathbb{R}^{p \times 1}$ is reshaped into a 2D image $C \in  \mathbb{R}^{p \times 1} \rightarrow C \in  \mathbb{R}^{h \times w}$ and fed into the classification network $F$. In order to obtain the pixel importance maps $P \in  \mathbb{R}^{h \times w}$ we take the isolated pixel image $X$ and the class label for the respective image $y$ and use the combination of $X,y$ to train the mixing layer with stochastic gradient descent. Subsequently, we utilize the corresponding mixing factors $\alpha$ of the analyzed image $I$ to obtain the pixel importance map $\alpha \in  \mathbb{R}^{p \times 1} \rightarrow P \in  \mathbb{R}^{h \times w}$.



    \subsection{Datasets}
    \label{sec:exp}
    PCIM is benchmarked on three high content imaging datasets NTR1, BBBC054 and BBBC010.

    \subsubsection{NTR1 Dataset}
    The data set represents an internalization assay targeting on neurotensin receptor 1 (NTR1). When activated, this G-protein-coupled receptor undergoes internalization into endosomes, facilitated by a beta-arrestin mediated process.
    Therefore, the activation of NTR1 is evaluated by measuring the redistribution of $\beta$-arrestin conjugated with green fluorescent protein (GFP) \cite{peddibhotla2013discovery}.\\
    The images of individual cells are defined by a nucleus detection following extraction of rectangular cell images of 112x112px size from the GFP fluorescence
    channel via commercial software \cite{steigele2020deep}.
    We use the two control states (non-internalized and internalized) of the assay as classes \cite{siegismund2022benchmarking} and ensure
    class balance through random under sampling \cite{JMLR:v18:16-365}.
    914 cell images are used, divided into a training set and a hold-out set, as a 80-20 split. The training set is further divided into an 80 percent training subset
    and a 20 percent validation subset.\\

    \subsubsection{BBBC054 Dataset}
    The dataset aims on characterizing Lipopolysaccharide (LPS)  activated immortalized mouse migroglia. Microglia are the resident immune cells of the central nervous system (CNS),
    including the brain and spinal cord playing a critical role in maintaining the health of the CNS \cite{he2021mouse}. LPS is a component of the outer membrane of certain bacteria. It’s recognized by the immune system as a signal of bacterial infection.
    The microglia become activated when they encounter LPS, and undergo phenotypic changes in their shape, function, and gene expression \cite{he2021mouse}.
    The dataset provides an annotated ground thruth
    of three different cell morphology classes: round, amoeboid and ramified\cite{ljosa2012annotated}.
    For our experiment we take the image from 2h exposure, cut out rectangular shaped annotated single cell images with a size of 32x32 px and ensure class balance through random under sampling \cite{JMLR:v18:16-365}.
    The data, which includes 1617 images afterwards, is partitioned into a training set and a hold-out set using an 80-20 ratio. The training set is then further split into two subsets: 80 percent for training and 20 percent for validation.

    \subsubsection{BBBC010}
    Dataset using \textit{Caenorhabditis elegans}, a  whole-animal model system, where the study aimed to identify small molecules with anti-infective
    properties in living organisms \cite{moy2009high}. The bacterium \textit{Enterococcus faecalis} was used to infect the \textit{C. elegans}. Some of these infected worms were left untreated,
    while others were treated with ampicillin, a known antibiotic for this pathogen.
    The untreated worms (negative control) primarily display a “dead” phenotype: they appear rod-like in shape and slightly uneven in texture.
    On the other hand, the treated worms (positive control, with ampicillin) primarily display a “alive” phenotype: they appear curved in shape and smooth in texture.\cite{ljosa2012annotated}.
    We use the fluorescence channel (Sytox marker) for network training and extract images of 320x320 pixels. This process results in 100 images.
    These images are then divided into a training set and a hold-out set at an 80-20 ratio. The training set is further split into two subsets,
    with 80 percent used for training and the remaining 20 percent for validation.\\
    A human annotated binary mask distinguishing between foreground (worms) and background is provided in addition to the original images.
    These masks are crucial for evaluating the ability of pixel attribution methods to localize the
    most significant pixels within the foreground. Among the three datasets used in this study, only the BBBC010 dataset provides this
    requisite mask. Thus, the assessment of localization ability is exclusively conducted for the BBBC010 dataset.

    \subsection{Baseline Methods}
    PCIM is benchmarked against five established baseline attribution methods:
    \begin{enumerate}
    	\item \textbf{Saliency maps \cite{simonyan2013deep}} compute the derivative of the class probability with respect to each pixel in the input image.
    	The magnitude of the gradient denotes which pixels need to be changed the least to affect the class score the most.
    	\item \textbf{RISE \cite{petsiuk2018rise}} produces attribution maps by probing the model with randomly masked versions of the input image and
    	observing the corresponding outputs. As black-box model, RISE doesn’t need any internal network states or gradients of the models.
    	\item \textbf{Grad-CAM (Gradient-weighted Class Activation Mapping) \cite{selvaraju2016grad}} uses the gradient of the predicted class score concerning the feature maps of the last convolutional layer.
    	\item \textbf{Grad-CAM++ \cite{chattopadhay2018grad}} in contrast uses a weighted combination of the positive partial derivatives of the last convolutional
    	layer feature maps. Both, Grad-CAM and Grad-CAM++ are the most commonly used attribution techniques, also in biomedical imaging \cite{van2022explainable}.
    	\item \textbf{Integrated Gradients \cite{sundararajan2017axiomatic}} involves creating a path from a baseline input to the actual input, which implies making predictions at each step, calculating gradients, and summing these gradients to compute the pixel importance.
    \end{enumerate}

    \subsection{Evaluation Metrics}
    \label{eval_metr}
    Two different categories of metrics are used, (1) to characterize the model fidelity and (2) to measure the localization ability of the attribution method. 
    \paragraph{Model Fidelity}
    For assessing model fidelity, we utilize the median Area Under the Curve (AUC) for Deletion and Insertion, as proposed by
    \cite{petsiuk2018rise}. The median AUC of Deletion measures a model's robustness by gradually removing pixels from the input image based on
    their importance. The model's classification performance of the image is tracked as pixels are deleted. The ideal curve decreases sharply with the deletion of the
    most important pixels (according to the calculated attribution maps) and remains low. From this deletion curve the AUC is calculated.
    The median AUC is then taken over all input images. The Median AUC of Insertion measures the opposite where important image pixels are incrementally
    added to a blank input, according to their importance.

    \paragraph{Localization Ability}
    For the localization metric we use the median relevance mass accuracy and the median relevance rank accuracy as introduced by \citep{arras2022clevr}.
    The median rank accuracy quantifies the proportion of the top ranked attributions that are contained within the ground truth mask, relative to the total
    size of the ground truth mask.
    In contrast, the median mass accuracy quantifies the proportion of positive attributions that are located within the ground truth mask, compared to the
    total number of positive attributions. For further details please refer to \cite{arras2022clevr}.

    \section{Results and Discussion}

    We here present a new method for calculating pixel attribution maps, which we call Pixel-wise Channel Isolation Mixing (PCIM).
    It uses trainable weights to mix image pixels for single-channel image classification tasks (refer to Figure \ref{fig:method_overview}
    and Section \ref{sec:methods}).
    The weights are used as pixel attribution values to denote the importance of each pixel in an image. We achieve state-of-the-art performance in
    terms of model fidelity and localization ability especially for fluorescence high content imaging (see Section \ref{sec:exp}).\\
    For model fidelity (which is measured by the median AUC of Deletion and Insertion - see section \ref{eval_metr}), PCIM performs exceptionally well.
    It surpasses the performance of all baseline methods in 83\% of the instances for the VGG16 downstream classification network (see Table \ref{combined_vgg16}; for results of the EfficientNetV2 as downstream classification network please see the Appendix section).\\
    Overall, the PCIM consistently shows state-of-the-art results in all tested datasets, but excels especially for the deletion task (Median AUC of deletion) in NTR1
    and BBBC054 datasets. We speculate that this is due to  PCIM’s strength for pinpointing features that, upon removal,
    considerably influence the model’s decision - a characteristic of the deletion task. Contrary,
    it slightly works less good at identifying a minimal set of features that can independently yield the same decision,
    a trait associated with the insertion task. The same behavior has been demonstrated, for instance, in the case of Integrated Gradients (Int. Grads) \cite{mahlau2022fidelity}.\\
    For BBBC010, we also computed the localization capability of the attribution methods (measured via median mass accuracy amd median rank accuracy),
    instead of just model fidelity (Table \ref{tab:bbbc10_vgg16_accuracy}). PCIM  demonstrates competitive state-of-the-art performance on both metrics, and stronger for
    median mass accuracy. In principle, the median mass accuracy quantifies how many of the positive attributions (important pixels identified) actually align with the true important pixels (as defined by the ground truth mask).
    PCIM surpasses all baseline methods in mass accuracy, with Integrated Gradients ranking second. This has also been shown previously with wide variety of attribution methods scoring higher mass accuracy values \cite{arras2022clevr}.


    \begin{table}[h]
        \centering
        \caption{Performance Metrics for different pixel attribution Methods on the NTR1, BBBC054 and BBBC010 data for VGG16. Model fidelity measured via Median Area under Curve (AUC) of Deletion and Insertion.}
        \label{combined_vgg16}
        \begin{tabular}{l@{}cccccc}
            \toprule
            & \multicolumn{2}{c}{\textbf{NTR1}} & \multicolumn{2}{c}{\textbf{BBBC054}} & \multicolumn{2}{c}{\textbf{BBBC010}} \\
            \cmidrule(lr){2-3} \cmidrule(lr){4-5} \cmidrule(lr){6-7}
            & \multicolumn{2}{c}{\textbf{Median AUC}} & \multicolumn{2}{c}{\textbf{Median AUC}} & \multicolumn{2}{c}{\textbf{Median AUC}} \\
            \textbf{Method} & \textbf{Deletion $\downarrow$} & \textbf{Insertion $\uparrow$} & \textbf{Deletion $\downarrow$} & \textbf{Insertion $\uparrow$} & \textbf{Deletion $\downarrow$} & \textbf{Insertion $\uparrow$}\\
            \midrule
            Saliency & 0.250 & 0.738 & 0.652 & \textbf{0.896} & 0.490 & 0.571 \\
            Rise & 0.261 & 0.739 & 0.652 & \textbf{0.896} & 0.479 & 0.571 \\
            Grad-CAM & 0.287 & 0.724 & 0.681 & 0.895 & 0.500 & 0.563 \\
            Grad-CAM++ & 0.299 & 0.726 & 0.705 & 0.894 & 0.523 & 0.564 \\
            Int. Grads & 0.258 & 0.739 & 0.655 & \textbf{0.896} & 0.470 & 0.573 \\
            \midrule
            \textbf{PCIM (ours)} & \textbf{0.112} & \textbf{0.889} & \textbf{0.199} & 0.866 & \textbf{0.464} & \textbf{0.637} \\
            \bottomrule
        \end{tabular}
    \end{table}

    \begin{table}[h!]
        \centering
        \caption{Localization performance measured as Median Accuracy for VGG16 on the BBBC010 data. \label{tab:bbbc10_vgg16_accuracy}}
        \begin{tabular}{cccc}
            \toprule
            & \multicolumn{2}{c}{\textbf{BBBC010}} \\
            \cmidrule(lr){2-3}
            &  \multicolumn{2}{c}{\textbf{Median Accuracy}} \\
            \textbf{Method} & \textbf{Rank $\uparrow$} & \textbf{Mass $\uparrow$} \\
            \midrule
            Saliency & 0.134 & 0.089 \\
            Rise & 0.243 & 0.143 \\
            Grad-CAM & 0.092 & 0.086 \\
            Grad-CAM++ & 0.088 & 0.086 \\
            Int. Grads & \textbf{0.617}  & 0.281 \\
            \midrule
            \textbf{PCIM (ours)} & 0.346 & \textbf{0.570} \\
            \bottomrule
        \end{tabular}
    \end{table}

    \subsection{Comparison of attribution methods}
    \begin{figure}[h!]
        \centering
        \includegraphics[width=0.8\linewidth]{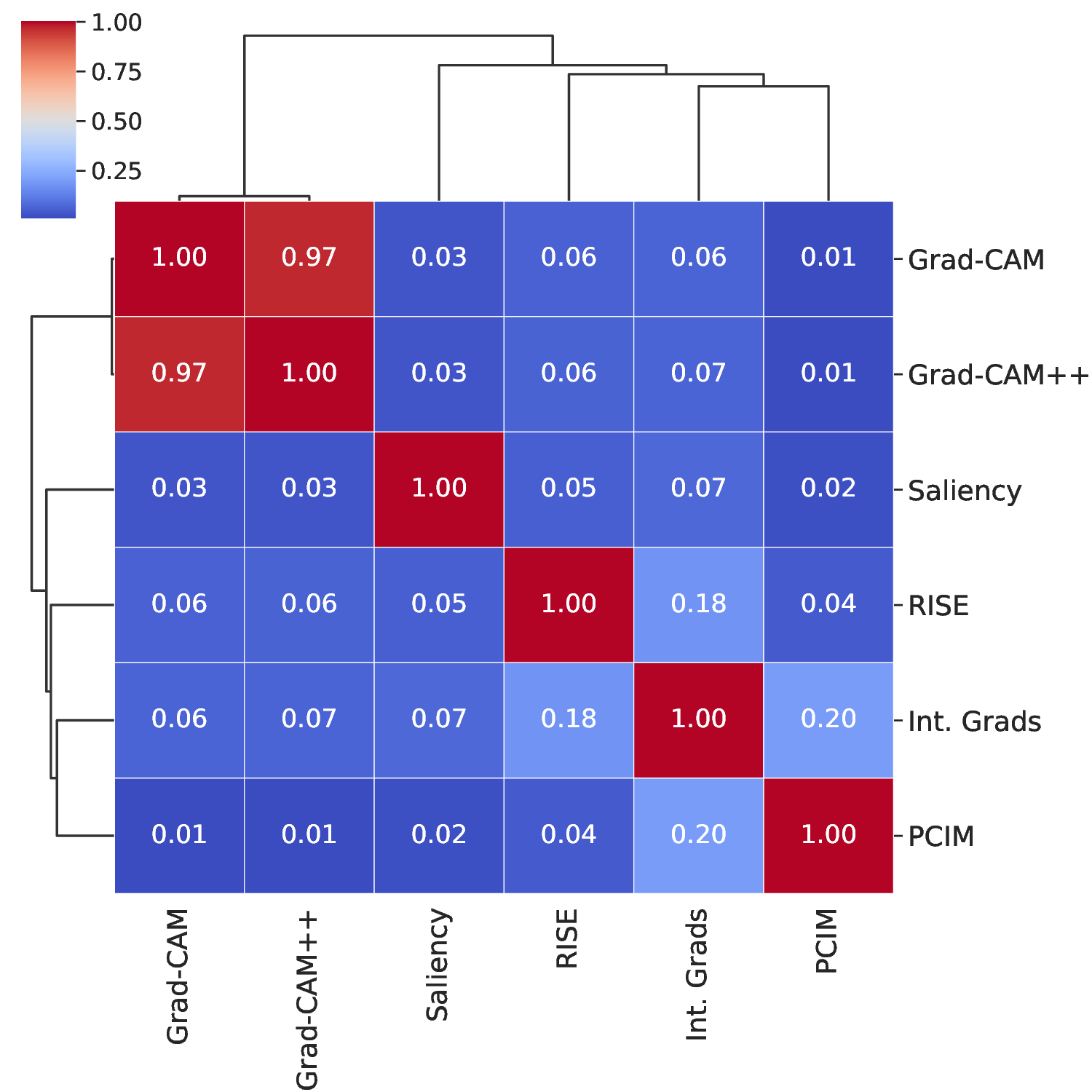}
        \caption{Cluster map showing the median Structural Similarity Index Measure (SSIM) \cite{wang2004image} for pixel attribute methods across all datasets (based on VGG16 network classification).
        The clustering uses Euclidean distance and average linkage. The color scale ranges from 0 to 1,
            where 1 indicates perfect agreement and 0 indicates no similarity.}

        \label{fig:cluster_ssim}
    \end{figure}
    The tested attribution methods are compared by the pairwise similarity of images using the Structural Similarity Index Measure (SSIM) \cite{wang2004image}. SSIM extracts three features from an image: luminance,
    contrast, and structure, and scores these features between the two images. Unlike methods that estimate
    absolute errors, SSIM reports on changes in structural information, thereby providing a more accurate representation of the perceived similarity. An analysis of the median
    SSIM \cite{wang2004image} of the attribution maps on all three datasets is shown in Figure \ref{fig:cluster_ssim}.\\ 
    Interestingly, the PCIM method is most similar to the Integrated Gradients (Int. Grads) method \cite{sundararajan2017axiomatic}. The IG
    method calculates the integral of the gradients between an input image and a baseline image, which is a black image.
    Likewise, PCIM uses a baseline image that is black (where the alpha values of the pixels are 0). After setting this baseline,
    PCIM then trains the alpha values for each pixel in the image, resulting in a pixel importance map that is subsequently used as pixel attribution map.
    Other attribution methods show significantly lower agreement with PCIM, with values less than 0.04.

    \subsection{Qualitative assessment of the attribution maps produced by PCIM}
    \label{qual_results}

    \begin{figure}[h!]
        \centering
        \includegraphics[width=1.0\linewidth]{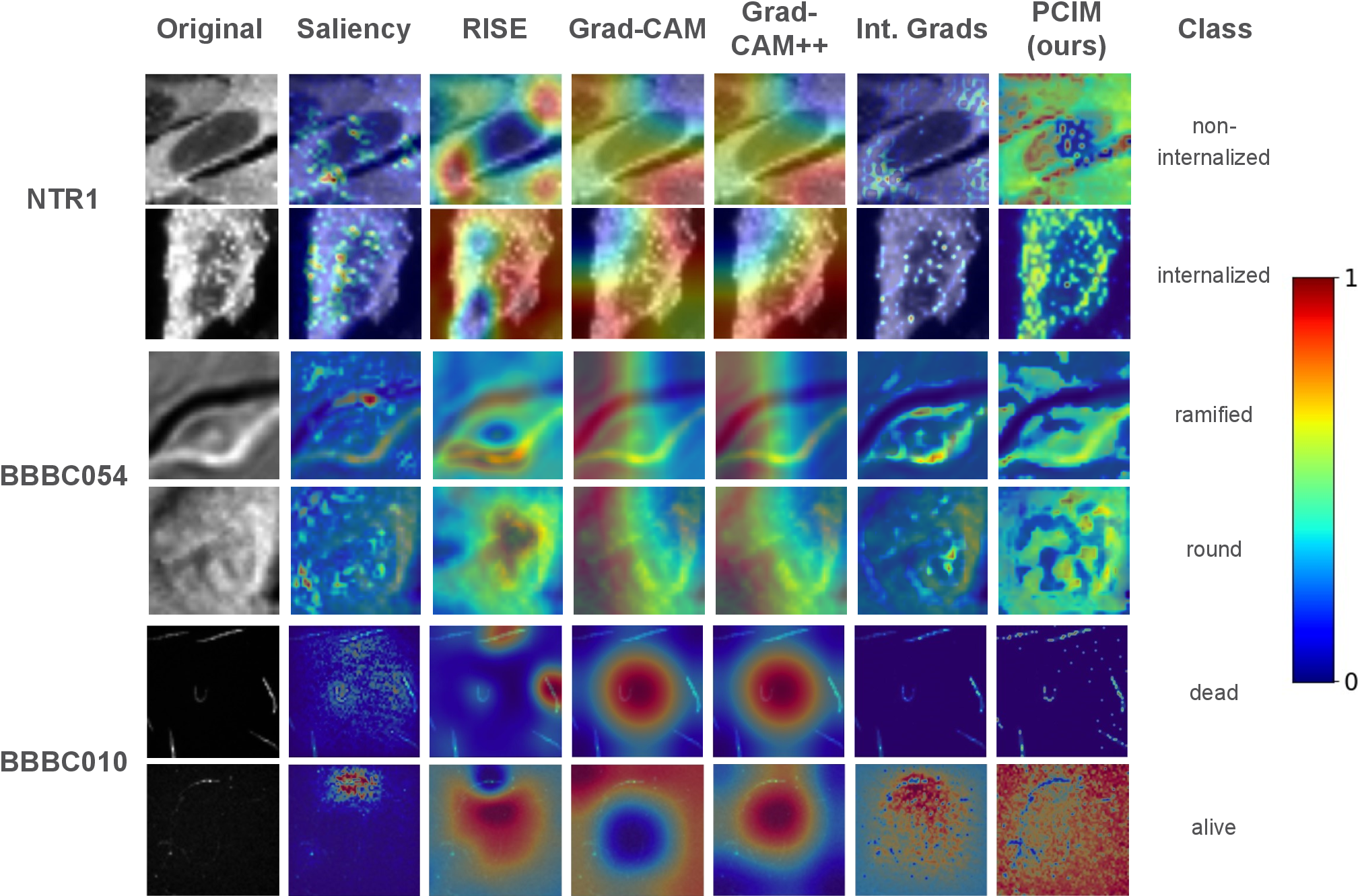}
        \caption{Example images per dataset (NTR1, BBBC054 and BBBC010) are shown per two lines, each column shows different
        pixel attribution methods were the produced maps are overlaid with the original images.
        The colorscale ranges from 0 to 1 where pixels of high importance are indicated in red (1), while those of low importance are shown in
        blue (0). Detailed explanation follows per dataset.     \underline{NTR1}: First row: Non-activated (non-internalized) phenotype: cells with spindle-shaped / polygonal morphology. The EGFP signal of $\beta$-arrestin is uniformly distributed. Second row: Activated (internalized): cells treated with neurotensin 1 peptide, showing the expected $\beta$-arrestin recruitment. The EGFP signal
        is primarily localized at the cell membrane and in the cytoplasm, reflecting the translocation of $\beta$-arrestin to the activated NTR1 receptors \cite{peddibhotla2013discovery}.
        \underline{BBBC054}: Third row: first non-activated phenotype (ramified); Fourth row: an activated phenotype (round) of LPS activated immortalized mouse migroglia. The activated migrolia possess a typical round/amoeboid
        morphology and texture \cite{woodburn2021semantics,sousa2017cellular} whereas the non-activated state cell possesses a smooth body \cite{leyh2021classification}.
        \underline{BBBC010}: Fifth row: 'dead' phenotype; Sixth row: 'alive' phenotype of \textit{C. elegans} in a live/dead assay.
        }

        \label{fig:attr_map_all}
    \end{figure}

    In support of interpreting high content imaging data, the qualitative assessment of pixel attribution maps can provide confidence into the validity of the obtained result and its interpretation.
    This is because not just the final end decision of a model is crucial, but also the process by which it arrives at that decision \cite{nguyen2021effectiveness}. Through the qualitative analysis of attribution maps, we gain the ability to visually interpret and comprehend the model’s decision-making process,
    to assess if the underlying biological hypothesis is reflected \cite{nguyen2021effectiveness}.
    Figure \ref{fig:attr_map_all} shows exemplary attribution maps for each baseline methods plus PCIM in all tested high content imaging datasets
    (NTR1, BBBC054, BBBC010).
    Overall, PCIM generates intricate maps that are comparable in resolution as those produced by Saliency or Integrated Gradients (Int. Grads) and contrasts
    with results from Grad-CAM(++) and RISE. These latter methods rely on data coming from a specific network layer or even just use broad masks, inherently
    resulting in maps at lower resolution.\\
    For the NTR1 dataset, the biological relevant image features are the differences in the intensity of EGFP fluorescence signal, localization and number of
    fluorescent dots in a cell \cite{peddibhotla2013discovery}. Using PCIM (see the NTR1 section in Figure \ref{fig:attr_map_all}) to interpret the network’s reasoning, it appears that the network focuses on the image areas of the ‘internalized’ phenotype with lower grey values, which are adjacent to brighter, widespread
    GFP signals for classification. These darker image parts of the ‘internalized’ phenotype are identified as having the highest importance.
    For the second ‘internalized’ class, PCIM clearly demonstrates the network using the true and expected ‘signal’. This involves recognizing the receptor’s internalization
    into the endosomes during classification, thereby reconstructing the biologically expected outcome. In contrast, Grad-CAM(++) and RISE (columns 3-5) provide
    only a displayof broad area of importance, which do not allow for any biological interpretation.\\
    The surface texture of microglia in the BBBC054 dataset changes as they become activated. Resting microglia have a smooth surface, while activated
    microglia exhibit a rougher texture due to the formation of membrane ruffles and increased phagocytotic activity \cite{campagno2021rapid}.
    hat has also functional implications. The rougher texture and increased organelle
    content are associated with the microglia’s enhanced ability to engulf and digest cellular debris, pathogens, and dead cells \cite{town2005microglial}.
    In the BBBC054 section of \ref{fig:attr_map_all}, it is shown that the PCIM can identify the aforementioned biologically important rough, protruding surface parts of the activated microglia cell as the most crucial for classification.
    Interestingly, for the non-activated state, PCIM identifies the attachment points of the cells' processes to the cell body as the most important image parts, which is indeed a relevant feature of
    unactivated, ramified microglia cells \cite{leyh2021classification}.
    The BBBC010 section of \ref{fig:attr_map_all} displays both classes in this dataset (dead/alive). The alive phenotype corresponds to worms treated with ampicillin and exhibit a low abundance of the Sytox marker,
    while the dead phenotype show a strong Sytox signal \cite{moy2009high}. PCIM identifies the high-intensity, folded, texture-rich parts of the cell membrane as the most important for classification, which is in line with the biology \cite{lively2013microglial}.
    Again, other methods, such as RISE and Grad-CAM(++), which perform well on natural images \cite{petsiuk2018rise}, lack the resolution capabilities necessary for biological understanding.
    Interestingly the trained network uses also dark background pixels as signal to distinguish between the 2 classes (see BBBC010 section in Figure \ref{fig:attr_map_all}).
    Using these features, the network delivers a perfect classification score on the hold out set (see Appendix Table \ref{tab:overview_tab}).
    Even if the classification performance is high, the network decision process may not reflect the biological interpretation, and it is important to distinguish the worms on the basis of their morphology (straight vs. roundish).
    PCIM  in particular demonstrates the significance of the Sytox signal in the classification of the ‘dead’ phenotype. Contrarily, it is the absence of the Sytox signal that is crucial for the ‘alive’ classification.
    Despite the presence of the Sytox signal within the ‘alive’ class, the pixels displaying this signal have a minimal impact on the classification decision (blue pixels in PCIM do overlap with Sytox signal in the original image).

    \section{Conclusion}
    We here introduce a new technique called Pixel-wise Channel Isolation Mixing (PCIM) for computing pixel attribution maps. Our findings demonstrate that
    PCIM reflects a unique approach contrasting to existing methods. Moreover, PCIM achieves state-of-the-art results for the biomedical
    image datasets in this study - which includes different types of imaging modalities like fluorescent and brightfield imaging.


    \bibliographystyle{unsrtnat}
    \bibliography{arxiv_pcim}  

\begin{thebibliography}{47}
\providecommand{\natexlab}[1]{#1}
\providecommand{\url}[1]{\texttt{#1}}
\expandafter\ifx\csname urlstyle\endcsname\relax
  \providecommand{\doi}[1]{doi: #1}\else
  \providecommand{\doi}{doi: \begingroup \urlstyle{rm}\Url}\fi

\bibitem[Kirillov et~al.(2023)Kirillov, Mintun, Ravi, Mao, Rolland, Gustafson,
  Xiao, Whitehead, Berg, Lo, Dollar, and Girshick]{Kirillov_2023_ICCV}
Alexander Kirillov, Eric Mintun, Nikhila Ravi, Hanzi Mao, Chloe Rolland, Laura
  Gustafson, Tete Xiao, Spencer Whitehead, Alexander~C. Berg, Wan-Yen Lo, Piotr
  Dollar, and Ross Girshick.
\newblock Segment anything.
\newblock In \emph{Proceedings of the IEEE/CVF International Conference on
  Computer Vision (ICCV)}, pages 4015--4026, October 2023.

\bibitem[Krizhevsky et~al.(2012)Krizhevsky, Sutskever, and
  Hinton]{NIPS2012_c399862d}
Alex Krizhevsky, Ilya Sutskever, and Geoffrey~E Hinton.
\newblock Imagenet classification with deep convolutional neural networks.
\newblock In F.~Pereira, C.J. Burges, L.~Bottou, and K.Q. Weinberger, editors,
  \emph{Advances in Neural Information Processing Systems}, volume~25. Curran
  Associates, Inc., 2012.

\bibitem[Lowe(2004)]{Lowe:2004:DIF:993451.996342}
David~G. Lowe.
\newblock Distinctive image features from scale-invariant keypoints.
\newblock \emph{Int. J. Comput. Vision}, 60:\penalty0 91--110, 2004.
\newblock ISSN 0920-5691.

\bibitem[Ribeiro et~al.(2016)Ribeiro, Singh, and Guestrin]{ribeiro2016should}
Marco~Tulio Ribeiro, Sameer Singh, and Carlos Guestrin.
\newblock " why should i trust you?" explaining the predictions of any
  classifier.
\newblock In \emph{Proceedings of the 22nd ACM SIGKDD international conference
  on knowledge discovery and data mining}, pages 1135--1144, 2016.

\bibitem[Lipton(2018)]{lipton2018mythos}
Zachary~C Lipton.
\newblock The mythos of model interpretability: In machine learning, the
  concept of interpretability is both important and slippery.
\newblock \emph{Queue}, 16\penalty0 (3):\penalty0 31--57, 2018.

\bibitem[Rao et~al.(2024)Rao, B{\"o}hle, and Schiele]{rao2024better}
Sukrut Rao, Moritz B{\"o}hle, and Bernt Schiele.
\newblock Better understanding differences in attribution methods via
  systematic evaluations.
\newblock \emph{IEEE Transactions on Pattern Analysis and Machine
  Intelligence}, 2024.

\bibitem[Nazir et~al.(2023)Nazir, Dickson, and Akram]{nazir2023survey}
Sajid Nazir, Diane~M Dickson, and Muhammad~Usman Akram.
\newblock Survey of explainable artificial intelligence techniques for
  biomedical imaging with deep neural networks.
\newblock \emph{Computers in Biology and Medicine}, page 106668, 2023.

\bibitem[Zhou et~al.(2016)Zhou, Khosla, Lapedriza, Oliva, and
  Torralba]{zhou2016learning}
Bolei Zhou, Aditya Khosla, Agata Lapedriza, Aude Oliva, and Antonio Torralba.
\newblock Learning deep features for discriminative localization.
\newblock In \emph{Proceedings of the IEEE conference on computer vision and
  pattern recognition}, pages 2921--2929, 2016.

\bibitem[Selvaraju et~al.(2016)Selvaraju, Das, Vedantam, Cogswell, Parikh, and
  Batra]{selvaraju2016grad}
Ramprasaath~R Selvaraju, Abhishek Das, Ramakrishna Vedantam, Michael Cogswell,
  Devi Parikh, and Dhruv Batra.
\newblock Grad-cam: Why did you say that?
\newblock \emph{arXiv preprint arXiv:1611.07450}, 2016.

\bibitem[Chattopadhay et~al.(2018)Chattopadhay, Sarkar, Howlader, and
  Balasubramanian]{chattopadhay2018grad}
Aditya Chattopadhay, Anirban Sarkar, Prantik Howlader, and Vineeth~N
  Balasubramanian.
\newblock Grad-cam++: Generalized gradient-based visual explanations for deep
  convolutional networks.
\newblock In \emph{2018 IEEE winter conference on applications of computer
  vision (WACV)}, pages 839--847. IEEE, 2018.

\bibitem[Ramaswamy et~al.(2020)]{ramaswamy2020ablation}
Harish~Guruprasad Ramaswamy et~al.
\newblock Ablation-cam: Visual explanations for deep convolutional network via
  gradient-free localization.
\newblock In \emph{proceedings of the IEEE/CVF winter conference on
  applications of computer vision}, pages 983--991, 2020.

\bibitem[Simonyan et~al.(2013)Simonyan, Vedaldi, and
  Zisserman]{simonyan2013deep}
Karen Simonyan, Andrea Vedaldi, and Andrew Zisserman.
\newblock Deep inside convolutional networks: Visualising image classification
  models and saliency maps.
\newblock \emph{arXiv preprint arXiv:1312.6034}, 2013.

\bibitem[Shrikumar et~al.(2017)Shrikumar, Greenside, and
  Kundaje]{shrikumar2017learning}
Avanti Shrikumar, Peyton Greenside, and Anshul Kundaje.
\newblock Learning important features through propagating activation
  differences.
\newblock In \emph{International conference on machine learning}, pages
  3145--3153. PMLR, 2017.

\bibitem[Sundararajan et~al.(2017)Sundararajan, Taly, and
  Yan]{sundararajan2017axiomatic}
Mukund Sundararajan, Ankur Taly, and Qiqi Yan.
\newblock Axiomatic attribution for deep networks.
\newblock In \emph{International conference on machine learning}, pages
  3319--3328. PMLR, 2017.

\bibitem[Petsiuk et~al.(2018)Petsiuk, Das, and Saenko]{petsiuk2018rise}
Vitali Petsiuk, Abir Das, and Kate Saenko.
\newblock Rise: Randomized input sampling for explanation of black-box models.
\newblock \emph{arXiv preprint arXiv:1806.07421}, 2018.

\bibitem[Fong and Vedaldi(2017)]{fong2017interpretable}
Ruth~C Fong and Andrea Vedaldi.
\newblock Interpretable explanations of black boxes by meaningful perturbation.
\newblock In \emph{Proceedings of the IEEE international conference on computer
  vision}, pages 3429--3437, 2017.

\bibitem[Van~der Velden et~al.(2022)Van~der Velden, Kuijf, Gilhuijs, and
  Viergever]{van2022explainable}
Bas~HM Van~der Velden, Hugo~J Kuijf, Kenneth~GA Gilhuijs, and Max~A Viergever.
\newblock Explainable artificial intelligence (xai) in deep learning-based
  medical image analysis.
\newblock \emph{Medical Image Analysis}, 79:\penalty0 102470, 2022.

\bibitem[R{\"o}hrl et~al.(2023)R{\"o}hrl, Maier, Lengl, Klenk, Heim, Knopp,
  Schumann, Hayden, and Diepold]{rohrl2023explainable}
Stefan R{\"o}hrl, Hendrik Maier, Manuel Lengl, Christian Klenk, Dominik Heim,
  Martin Knopp, Simon Schumann, Oliver Hayden, and Klaus Diepold.
\newblock Explainable artificial intelligence for cytological image analysis.
\newblock In \emph{International Conference on Artificial Intelligence in
  Medicine}, pages 75--85. Springer, 2023.

\bibitem[Buchser et~al.(2014)Buchser, Collins, Garyantes, Guha, Haney, Lemmon,
  Li, and Trask]{buchser2014assay}
William Buchser, Mark Collins, Tina Garyantes, Rajarshi Guha, Steven Haney,
  Vance Lemmon, Zhuyin Li, and O~Joseph Trask.
\newblock Assay development guidelines for image-based high content screening,
  high content analysis and high content imaging.
\newblock \emph{Assay guidance manual [Internet]}, 2014.

\bibitem[Usaj et~al.(2016)Usaj, Styles, Verster, Friesen, Boone, and
  Andrews]{usaj2016high}
Mojca~Mattiazzi Usaj, Erin~B Styles, Adrian~J Verster, Helena Friesen, Charles
  Boone, and Brenda~J Andrews.
\newblock High-content screening for quantitative cell biology.
\newblock \emph{Trends in cell biology}, 26\penalty0 (8):\penalty0 598--611,
  2016.

\bibitem[Siegismund et~al.(2024)Siegismund, Wieser, Heyse, and
  Steigele]{siegismund2023learning}
Daniel Siegismund, Mario Wieser, Stephan Heyse, and Stephan Steigele.
\newblock Learning channel importance for high content imaging with
  interpretable deep input channel mixing.
\newblock In \emph{Pattern Recognition. DAGM GCPR 2023}, page 335–347.
  Springer Nature Switzerland, 2024.
\newblock ISBN 9783031546051.
\newblock \doi{10.1007/978-3-031-54605-1_22}.
\newblock URL \url{http://dx.doi.org/10.1007/978-3-031-54605-1_22}.

\bibitem[Simonyan and Zisserman(2014)]{simonyan2014very}
Karen Simonyan and Andrew Zisserman.
\newblock Very deep convolutional networks for large-scale image recognition.
\newblock \emph{arXiv preprint arXiv:1409.1556}, 2014.

\bibitem[Tan and Le(2021)]{tan2021efficientnetv2}
Mingxing Tan and Quoc Le.
\newblock Efficientnetv2: Smaller models and faster training.
\newblock In \emph{International conference on machine learning}, pages
  10096--10106. PMLR, 2021.

\bibitem[Peddibhotla et~al.(2013)Peddibhotla, Hedrick, Hershberger, Maloney,
  Li, Milewski, Gosalia, Gray, Mehta, Sugarman,
  et~al.]{peddibhotla2013discovery}
Satyamaheshwar Peddibhotla, Michael~P Hedrick, Paul Hershberger, Patrick~R
  Maloney, Yujie Li, Monika Milewski, Palak Gosalia, Wilson Gray, Alka Mehta,
  Eliot Sugarman, et~al.
\newblock Discovery of ml314, a brain penetrant nonpeptidic $\beta$-arrestin
  biased agonist of the neurotensin ntr1 receptor.
\newblock \emph{ACS medicinal chemistry letters}, 4\penalty0 (9):\penalty0
  846--851, 2013.

\bibitem[Ljosa et~al.(2012)Ljosa, Sokolnicki, and
  Carpenter]{ljosa2012annotated}
Vebjorn Ljosa, Katherine~L Sokolnicki, and Anne~E Carpenter.
\newblock Annotated high-throughput microscopy image sets for validation.
\newblock \emph{Nature Methods}, 9\penalty0 (7):\penalty0 637--637, 2012.

\bibitem[Zhang et~al.(2020)Zhang, Wen, and Shi]{zhang_deep_2020}
Lingzhi Zhang, Tarmily Wen, and Jianbo Shi.
\newblock Deep {Image} {Blending}.
\newblock pages 231--240. IEEE Computer Society, March 2020.
\newblock ISBN 978-1-72816-553-0.
\newblock \doi{10.1109/WACV45572.2020.9093632}.
\newblock URL
  \url{https://www.computer.org/csdl/proceedings-article/wacv/2020/09093632/1jPbdjek65i}.

\bibitem[Kokalj and Somrak(2019)]{kokalj_why_2019}
Žiga Kokalj and Maja Somrak.
\newblock Why {Not} a {Single} {Image}? {Combining} {Visualizations} to
  {Facilitate} {Fieldwork} and {On}-{Screen} {Mapping}.
\newblock \emph{Remote Sensing}, 11\penalty0 (7):\penalty0 747, January 2019.
\newblock ISSN 2072-4292.
\newblock \doi{10.3390/rs11070747}.
\newblock URL \url{https://www.mdpi.com/2072-4292/11/7/747}.
\newblock Number: 7 Publisher: Multidisciplinary Digital Publishing Institute.

\bibitem[Steigele et~al.(2020)Steigele, Siegismund, Fassler, Kustec, Kappler,
  Hasaka, Yee, Brodte, and Heyse]{steigele2020deep}
Stephan Steigele, Daniel Siegismund, Matthias Fassler, Marusa Kustec, Bernd
  Kappler, Tom Hasaka, Ada Yee, Annette Brodte, and Stephan Heyse.
\newblock Deep learning-based hcs image analysis for the enterprise.
\newblock \emph{SLAS DISCOVERY: Advancing the Science of Drug Discovery},
  25\penalty0 (7):\penalty0 812--821, 2020.

\bibitem[Siegismund et~al.(2022)Siegismund, Fassler, Heyse, and
  Steigele]{siegismund2022benchmarking}
Daniel Siegismund, Matthias Fassler, Stephan Heyse, and Stephan Steigele.
\newblock Benchmarking feature selection methods for compressing image
  information in high-content screening.
\newblock \emph{SLAS technology}, 27\penalty0 (1):\penalty0 85--93, 2022.

\bibitem[Lema{{\^i}}tre et~al.(2017)Lema{{\^i}}tre, Nogueira, and
  Aridas]{JMLR:v18:16-365}
Guillaume Lema{{\^i}}tre, Fernando Nogueira, and Christos~K. Aridas.
\newblock Imbalanced-learn: A python toolbox to tackle the curse of imbalanced
  datasets in machine learning.
\newblock \emph{Journal of Machine Learning Research}, 18\penalty0
  (17):\penalty0 1--5, 2017.
\newblock URL \url{http://jmlr.org/papers/v18/16-365.html}.

\bibitem[He et~al.(2021)He, Taylor, Yao, and Bhattacharya]{he2021mouse}
Yingbo He, Natalie Taylor, Xiang Yao, and Anindya Bhattacharya.
\newblock Mouse primary microglia respond differently to lps and poly (i: C) in
  vitro.
\newblock \emph{Scientific Reports}, 11\penalty0 (1):\penalty0 10447, 2021.

\bibitem[Moy et~al.(2009)Moy, Conery, Larkins-Ford, Wu, Mazitschek, Casadei,
  Lewis, Carpenter, and Ausubel]{moy2009high}
Terence~I Moy, Annie~L Conery, Jonah Larkins-Ford, Gang Wu, Ralph Mazitschek,
  Gabriele Casadei, Kim Lewis, Anne~E Carpenter, and Frederick~M Ausubel.
\newblock High-throughput screen for novel antimicrobials using a whole animal
  infection model.
\newblock \emph{ACS chemical biology}, 4\penalty0 (7):\penalty0 527--533, 2009.

\bibitem[Arras et~al.(2022)Arras, Osman, and Samek]{arras2022clevr}
Leila Arras, Ahmed Osman, and Wojciech Samek.
\newblock Clevr-xai: A benchmark dataset for the ground truth evaluation of
  neural network explanations.
\newblock \emph{Information Fusion}, 81:\penalty0 14--40, 2022.

\bibitem[Mahlau and Nolde(2022)]{mahlau2022fidelity}
Yannik Mahlau and Christian Nolde.
\newblock Fidelity of ensemble aggregation for saliency map explanations using
  bayesian optimization techniques.
\newblock \emph{arXiv preprint arXiv:2207.01565}, 2022.

\bibitem[Wang et~al.(2004)Wang, Bovik, Sheikh, and Simoncelli]{wang2004image}
Zhou Wang, Alan~C Bovik, Hamid~R Sheikh, and Eero~P Simoncelli.
\newblock Image quality assessment: from error visibility to structural
  similarity.
\newblock \emph{IEEE transactions on image processing}, 13\penalty0
  (4):\penalty0 600--612, 2004.

\bibitem[Woodburn et~al.(2021)Woodburn, Bollinger, and
  Wohleb]{woodburn2021semantics}
Samuel~C Woodburn, Justin~L Bollinger, and Eric~S Wohleb.
\newblock The semantics of microglia activation: neuroinflammation,
  homeostasis, and stress.
\newblock \emph{Journal of neuroinflammation}, 18:\penalty0 1--16, 2021.

\bibitem[Sousa et~al.(2017)Sousa, Biber, and Michelucci]{sousa2017cellular}
Carole Sousa, Knut Biber, and Alessandro Michelucci.
\newblock Cellular and molecular characterization of microglia: a unique immune
  cell population.
\newblock \emph{Frontiers in immunology}, 8:\penalty0 245768, 2017.

\bibitem[Leyh et~al.(2021)Leyh, Paeschke, Mages, Michalski, Nowicki, Bechmann,
  and Winter]{leyh2021classification}
Judith Leyh, Sabine Paeschke, Bianca Mages, Dominik Michalski, Marcin Nowicki,
  Ingo Bechmann, and Karsten Winter.
\newblock Classification of microglial morphological phenotypes using machine
  learning.
\newblock \emph{Frontiers in cellular neuroscience}, 15:\penalty0 701673, 2021.

\bibitem[Nguyen et~al.(2021)Nguyen, Kim, and Nguyen]{nguyen2021effectiveness}
Giang Nguyen, Daeyoung Kim, and Anh Nguyen.
\newblock The effectiveness of feature attribution methods and its correlation
  with automatic evaluation scores.
\newblock \emph{Advances in Neural Information Processing Systems},
  34:\penalty0 26422--26436, 2021.

\bibitem[Campagno et~al.(2021)Campagno, Lu, Jassim, Albalawi, Cenaj, Tso,
  Clark, Sripinun, G{\'o}mez, and Mitchell]{campagno2021rapid}
Keith~E Campagno, Wennan Lu, Assraa~Hassan Jassim, Farraj Albalawi, Aurora
  Cenaj, Huen-Yee Tso, Sophia~P Clark, Puttipong Sripinun, N{\'e}stor~M{\'a}s
  G{\'o}mez, and Claire~H Mitchell.
\newblock Rapid morphologic changes to microglial cells and upregulation of
  mixed microglial activation state markers induced by p2x7 receptor
  stimulation and increased intraocular pressure.
\newblock \emph{Journal of neuroinflammation}, 18:\penalty0 1--18, 2021.

\bibitem[Town et~al.(2005)Town, Nikolic, and Tan]{town2005microglial}
Terrence Town, Veljko Nikolic, and Jun Tan.
\newblock The microglial" activation" continuum: from innate to adaptive
  responses.
\newblock \emph{Journal of neuroinflammation}, 2:\penalty0 1--10, 2005.

\bibitem[Lively and Schlichter(2013)]{lively2013microglial}
Starlee Lively and Lyanne~C Schlichter.
\newblock The microglial activation state regulates migration and roles of
  matrix-dissolving enzymes for invasion.
\newblock \emph{Journal of neuroinflammation}, 10:\penalty0 1--14, 2013.

\bibitem[Ruder(2016)]{ruder2016overview}
Sebastian Ruder.
\newblock An overview of gradient descent optimization algorithms.
\newblock \emph{arXiv preprint arXiv:1609.04747}, 2016.

\bibitem[Leondgarse(2022)]{leondgarse}
Leondgarse.
\newblock Keras cv attention models.
\newblock \url{https://github.com/leondgarse/keras_cv_attention_models}, 2022.

\bibitem[Fel et~al.(2022)Fel, Hervier, Vigouroux, Poche, Plakoo, Cadene,
  Chalvidal, Colin, Boissin, Bethune, Picard, Nicodeme, Gardes, Flandin, and
  Serre]{fel2022xplique}
Thomas Fel, Lucas Hervier, David Vigouroux, Antonin Poche, Justin Plakoo, Remi
  Cadene, Mathieu Chalvidal, Julien Colin, Thibaut Boissin, Louis Bethune,
  Agustin Picard, Claire Nicodeme, Laurent Gardes, Gregory Flandin, and Thomas
  Serre.
\newblock Xplique: A deep learning explainability toolbox.
\newblock \emph{Workshop on Explainable Artificial Intelligence for Computer
  Vision (CVPR)}, 2022.

\bibitem[Abadi et~al.(2015)Abadi, Agarwal, Barham, Brevdo, Chen, Citro,
  Corrado, Davis, Dean, Devin, Ghemawat, Goodfellow, Harp, Irving, Isard, Jia,
  Jozefowicz, Kaiser, Kudlur, Levenberg, Man\'{e}, Monga, Moore, Murray, Olah,
  Schuster, Shlens, Steiner, Sutskever, Talwar, Tucker, Vanhoucke, Vasudevan,
  Vi\'{e}gas, Vinyals, Warden, Wattenberg, Wicke, Yu, and
  Zheng]{tensorflow2015-whitepaper}
Mart\'{i}n Abadi, Ashish Agarwal, Paul Barham, Eugene Brevdo, Zhifeng Chen,
  Craig Citro, Greg~S. Corrado, Andy Davis, Jeffrey Dean, Matthieu Devin,
  Sanjay Ghemawat, Ian Goodfellow, Andrew Harp, Geoffrey Irving, Michael Isard,
  Yangqing Jia, Rafal Jozefowicz, Lukasz Kaiser, Manjunath Kudlur, Josh
  Levenberg, Dandelion Man\'{e}, Rajat Monga, Sherry Moore, Derek Murray, Chris
  Olah, Mike Schuster, Jonathon Shlens, Benoit Steiner, Ilya Sutskever, Kunal
  Talwar, Paul Tucker, Vincent Vanhoucke, Vijay Vasudevan, Fernanda Vi\'{e}gas,
  Oriol Vinyals, Pete Warden, Martin Wattenberg, Martin Wicke, Yuan Yu, and
  Xiaoqiang Zheng.
\newblock {TensorFlow}: Large-scale machine learning on heterogeneous systems,
  2015.
\newblock URL \url{https://www.tensorflow.org/}.
\newblock Software available from tensorflow.org.

\bibitem[Hedstr{\"{o}}m et~al.(2023)Hedstr{\"{o}}m, Weber, Krakowczyk, Bareeva,
  Motzkus, Samek, Lapuschkin, and H{\"{o}}hne]{hedstrom2023quantus}
Anna Hedstr{\"{o}}m, Leander Weber, Daniel Krakowczyk, Dilyara Bareeva, Franz
  Motzkus, Wojciech Samek, Sebastian Lapuschkin, and Marina Marina~M.{-}C.
  H{\"{o}}hne.
\newblock Quantus: An explainable ai toolkit for responsible evaluation of
  neural network explanations and beyond.
\newblock \emph{Journal of Machine Learning Research}, 24\penalty0
  (34):\penalty0 1--11, 2023.
\newblock URL \url{http://jmlr.org/papers/v24/22-0142.html}.

\end{thebibliography}

    \section{Appendix}

    \subsection{Base Model Description}
     \subsubsection{VGG 16}
    We utilize the well-known VGG16 model structure, as outlined in \cite{simonyan2014very}. Our model takes one input channel and is trained
    from scratch for each task over 300 epochs. The training process uses stochastic gradient descent \cite{ruder2016overview}, which includes momentum and learning rate decay.
    The model weights that we use are from the training epoch that resulted in the lowest cross-entropy loss on the validation data.

    \subsubsection{EfficientNetV2}
    As second model we employ EfficientNetV2B0 \cite{tan2021efficientnetv2}. Again only one input channel is used and the model is trained from scratch for each task over 300 epochs.
    The training procedure employs stochastic gradient descent, as described in \cite{ruder2016overview}, incorporating  momentum and learning rate decay.
    The weights of the model that we utilize are derived from the training epoch that yielded the minimum cross-entropy loss on the validation dataset. We apply the tensorflow keras implementation
    provided in \cite{leondgarse}.

    \subsection{Classification Performance}

        \begin{table}[ht]
    	\caption{Performance Metrics on the hold out set for the different Datasets and Networks}
    	\label{tab:overview_tab}
    	\begin{tabular}{lcccccccc}
    		\hline
    		\textbf{Dataset} & \textbf{Classes} & \textbf{Image Size} & \textbf{Hold-Out Set Size} & \textbf{Network} & \textbf{Accuracy} & \textbf{Precision} & \textbf{Recall} & \textbf{F1} \\
    		\hline
    		NTR1 & 2 & 112px & 184 & VGG16 & 0.89 & 0.89 & 0.89 & 0.89 \\
    		& & & & EfficientNet & 0.83 & 0.83 & 0.82 & 0.82 \\
    		\hline
    		BBBC054 & 3 & 32px & 324 & VGG16 & 0.79 & 0.79 & 0.79 & 0.79 \\
    		& & & & EfficientNet & 0.74 & 0.75 & 0.74 & 0.74 \\
    		\hline
    		BBBC010 & 2 & 320px & 20 & VGG16 & 1.00 & 1.00 & 1.00 & 1.00 \\
    		& & & & EfficientNet & 0.93 & 0.94 & 0.93 & 0.93 \\
    		\hline
    	\end{tabular}
    \end{table}


    \paragraph{NTR1:} The accuracy of the models on the hold-out set ranges from 0.89 (for VGG16) to 0.83 (for EfficientNetV2B0), as shown in Table \ref{tab:overview_tab}.

    \paragraph{BBBC054:} VGG16 achieves an accuracy of 0.79 whereas EfficienNet is able to classify the hold-out set with an accuracy of 0.73 (Table \ref{tab:overview_tab}).

    \paragraph{BBBC010:} This approach has enabled us to achieve a perfect classification accuracy of 1.0 for the hold-out set using VGG16.
    However, when using EfficientNet, the classification accuracy is slightly lower at 0.93 (Table \ref{tab:overview_tab}).\\
    The provided segmentation masks enable to benchmark the localization performance of all attribution methods in addition to the Model Fidelity.

    \section{Implementation of Evaluation Metrics}
    \paragraph{Baseline Methods}
    We use the implementation provided from Xplique \cite{fel2022xplique} to calculate the attribution maps for the baselines.
    \paragraph{Model Fidelity}
        We've updated the original code from \cite{petsiuk2018rise} to work with Tensorflow 2 \cite{tensorflow2015-whitepaper}. For a more detailed explanation of both metrics,
    please refer to \cite{petsiuk2018rise}.

    \paragraph{Localization Ability}
    We utilize the implementation offered by Quantus, as detailed in \cite{hedstrom2023quantus}.

    \section{Results for EfficientNet}

    \begin{table}[h]
        \centering
        \caption{Performance Metrics for different pixel attribution Methods on the NTR1, BBBC054 and BBBC010 data for EfficientNet Network. Model fidelity measured via Median Area under Curve (AUC) of Deletion and Insertion.}
        \label{combined_effnet_all}
        \begin{tabular}{l@{}cccccc}
            \toprule
            & \multicolumn{2}{c}{\textbf{NTR1}} & \multicolumn{2}{c}{\textbf{BBBC054}} & \multicolumn{2}{c}{\textbf{BBBC010}} \\
            \cmidrule(lr){2-3} \cmidrule(lr){4-5} \cmidrule(lr){6-7}
            & \multicolumn{2}{c}{\textbf{Median AUC}} & \multicolumn{2}{c}{\textbf{Median AUC}} & \multicolumn{2}{c}{\textbf{Median AUC}} \\
            \textbf{Method} & \textbf{Deletion $\downarrow$} & \textbf{Insertion $\uparrow$} & \textbf{Deletion $\downarrow$} & \textbf{Insertion $\uparrow$} & \textbf{Deletion $\downarrow$} & \textbf{Insertion $\uparrow$}\\
            \midrule
            Saliency & 0.406 & 0.633 & 0.5592 & \textbf{0.863} & 0.143 & \textbf{1.000} \\
            Rise & 0.404 & 0.631 & 0.561 & \textbf{0.863} & 0.076 & \textbf{1.000} \\
            GradCAM & 0.407 & 0.630 & 0.575 & \textbf{0.863} & 0.223 & \textbf{1.000} \\
            GradCAM++ & 0.407 & 0.630 & 0.575 & \textbf{0.863} & 0.193 & \textbf{1.000} \\
            Int. Grads & 0.420 & 0.631 & 0.559 & \textbf{0.863} & \textbf{0.074} & \textbf{1.000} \\
            \midrule
            \textbf{PCIM (ours)} & \textbf{0.312} & \textbf{0.764} & \textbf{0.261} & 0.755 & 0.128 & \textbf{1.000} \\
            \bottomrule
        \end{tabular}
    \end{table}

    \begin{table}[h]
        \centering
        \caption{Localization performance measured as Median Accuracy for EfficientNet on the BBBC010 data.}\label{tab:bbbc10_effnet_accuracy}
        \begin{tabular}{cccc}
            \toprule
            &  \multicolumn{2}{c}{\textbf{Median Accuracy}} \\
            \textbf{Method} & \textbf{Rank $\uparrow$} & \textbf{Mass $\uparrow$} \\
            \midrule
            Saliency & 0.052 & 0.072 \\
            Rise & 0.180 & 0.092 \\
            GradCAM & 0.051 & 0.079 \\
            GradCAM++ & 0.060 & 0.081 \\
            IntegratedGradients & \textbf{0.381} & 0.091 \\
            \midrule
            \textbf{PCIM (ours)} & 0.225 & \textbf{0.121} \\
            \bottomrule
        \end{tabular}
    \end{table}

\end{document}